# Facial Expression Recognition with YOLOv11 and YOLOv12: A Comparative Study

1st Umma Aymon
*Faculty of Computing*
*Universiti Malaysia Pahang*
*Al-Sultan Abdullah*
Pahang, Malaysia
ummaaymon@gmail.com

2nd Nur Shazwani Kamarudin
*Faculty of Computing*
*Universiti Malaysia Pahang*
*Al-Sultan Abdullah*
Pahang, Malaysia
nshazwani@umpsa.edu.my*

3rd Ahmad Fakhri Ab. Nasir
*Faculty of Computing*
*Universiti Malaysia Pahang*
*Al-Sultan Abdullah*
Pahang, Malaysia
afakhri@umpsa.edu.my

*Abstract*—Facial Expression Recognition (FER) remains a challenging task, especially in unconstrained, real-world environments. This study investigates the performance of two lightweight models, YOLOv11n and YOLOv12n, which are the nano variants of the latest official YOLO series, within a unified detection and classification framework for FER. Two benchmark classification datasets, FER2013 (in the wild) and KDEF (studio quality), are converted into object detection format and model performance is evaluated using mAP@0.5, precision, recall, and confusion matrices. Results show that YOLOv12n achieves the highest overall performance on the clean KDEF dataset with a mAP@0.5 of 95.6%, and also outperforms YOLOv11n on the FER2013 dataset in terms of mAP (63.8%), reflecting stronger sensitivity to varied expressions. In contrast, YOLOv11n demonstrates higher precision (65.2%) on FER2013, indicating fewer false positives and better reliability in noisy, real-world conditions. On FER2013, both models show more confusion between visually similar expressions, while clearer class separation is observed on the cleaner KDEF dataset. These findings underscore the trade-off between sensitivity and precision, illustrating how lightweight YOLO models can effectively balance performance and efficiency. The results demonstrate adaptability across both controlled and real-world conditions, establishing these models as strong candidates for real-time, resource-constrained emotion-aware AI applications.

## I. INTRODUCTION

Facial Expression Recognition (FER) is the computational process of automatically identifying human emotions from facial images or videos by analyzing facial muscle movements and expressions. As a core task in affective computing, FER enables machines to interpret non-verbal cues, enhancing interaction in fields such as education, healthcare, surveillance, and driver monitoring. Despite growing adoption, FER still encounters significant difficulties in real-world scenarios due to variations in pose, lighting, occlusion, and background clutter [1], [2].

Traditional FER systems typically follow a two-stage pipeline: face detection followed by expression classification. While effective in controlled environments, this approach increases computational complexity and risks error propagation between stages. Deep learning, particularly Convolutional Neural Networks (CNNs), has improved performance, but many models still depend on cropped and aligned faces, limiting robustness in unconstrained settings [3].

To address these challenges, this study explores real-time object detection models that perform facial expression localization and classification in a single stage. Unlike traditional pipelines that separate face detection from emotion recognition, this unified approach reduces complexity and latency, making it more suitable for dynamic, real-world environments [4]. YOLO (You Only Look Once) models follow a single-shot architecture, comprising a backbone for feature extraction, a neck for multi-scale feature fusion, and a head for prediction. In this work, the official lightweight nano versions YOLOv11n [5] and YOLOv12n [6] are adapted from the Ultralytics YOLO series, optimized for real-time performance in environments with limited computational resources. YOLOv11 introduces the C3k2 module and Spatial Pyramid Pooling-Fast (SPPF) to improve spatial encoding and multi-scale aggregation, while YOLOv12 incorporates attention mechanisms and Residual Efficient Layer Aggregation Networks (R-ELAN) to enhance feature focus and representation [7]. These enhancements improve the model's ability to focus on subtle facial features, as illustrated in Figure 1, which shows the core architectural components of YOLOv11 and YOLOv12.

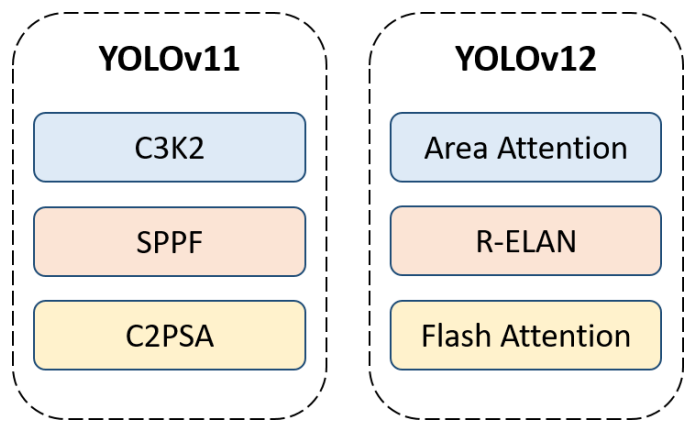

Fig. 1. Core architectural components distinguishing YOLOv11 and YOLOv12 models.

To evaluate model performance, two benchmark datasets are used: FER2013 [8] and KDEF [9]. FER2013 contains low-resolution, in-the-wild images, while KDEF offers high-resolution, studio-captured expressions. Both datasets were adapted to object detection format using a unified labeling strategy, ensuring consistent evaluation across diverse conditions.

While YOLOv5 and YOLOv8 have gained popularity in FER research, newer versions such as YOLOv11 and YOLOv12 remain underexplored [1]. Prior studies often focus on individual YOLO variants or limited datasets, lacking a comprehensive comparison across architectures.

This paper contributes the following:
- Adapts YOLOv11n and YOLOv12n for FER by converting classification datasets into object detection format.
- Evaluates both models on FER2013 and KDEF using unified preprocessing and training pipelines.
- Reports performance using mAP@0.5, precision, recall, and confusion matrices for class-wise insights.
- Demonstrates the potential of lightweight, detection oriented models for FER in resource constrained settings.

The remainder of the paper is organized as follows: Section II reviews related work. Section III details the datasets, models, and training pipeline. Section IV presents experimental setup and discusses results, and Section V concludes the paper.

## II. RELATED WORK

Facial Expression Recognition (FER) is a core component of affective computing that enables machines to interpret human emotions by analyzing facial cues. It plays a significant role across domains such as education, healthcare, automotive safety, surveillance, and human–robot interaction, supporting tasks like adaptive learning, mood assessment, fatigue detection, and emotion-aware engagement. Over time, FER has evolved from early rule-based systems to modern deep learning approaches, reflecting its growing importance in building emotionally intelligent systems [10].

### A. Classification-based FER

The dominant paradigm in facial expression recognition has traditionally treated the task as an image classification problem. In this approach, facial images, typically cropped and aligned, are passed through a model that assigns an expression label based on learned visual features.

Early classification-based FER systems relied on hand-crafted features such as Local Binary Patterns (LBP), Histogram of Oriented Gradients (HOG), and Gabor filters. These features were often combined with classical classifiers like Support Vector Machines (SVM) or k-Nearest Neighbors (k-NN) to identify facial expressions [1]. While interpretable and lightweight, these methods lacked robustness in unconstrained environments due to their limited capacity to model complex variations in appearance.

With the advent of deep learning, Convolutional Neural Networks (CNNs) became the foundation for modern FER systems. Architectures such as VGGNet, ResNet, and MobileNet have demonstrated strong performance by learning hierarchical feature representations directly from data [10]. More recently, Vision Transformers (ViTs) and hybrid CNN-Transformer models have been explored to capture global dependencies and improve performance in challenging scenarios [11].

Despite these advances, classification-based FER models remain sensitive to occlusions, pose variations, and background clutter. Moreover, these models typically require accurate face detection and alignment as a preprocessing step, which can introduce additional complexity and performance bottlenecks. These limitations motivate the exploration of alternative, more integrated approaches such as object detection frameworks.

### B. Object-detection based FER

Object detection, a fundamental task in computer vision, involves not only identifying the presence of objects in an image or video but also locating them by providing a bounding box around each instance. This differs from image classification, which focuses solely on assigning a label to the entire image.

In the context of Facial Expression Recognition (FER), object detection models, particularly the YOLO (You Only Look Once) series, have gained popularity for directly locating and classifying facial expressions due to their real-time performance and architectural simplicity.

Recent studies have adapted various YOLO versions for FER tasks. Xu et al. [3] proposed a YOLOv8-based method that uses efficient face detection and convolutional networks to classify expressions, showing improved performance in complex environments with occlusion and lighting variations. Similarly, the FER-YOLO-Mamba model integrated YOLOv8 with a selective state-space mechanism to enhance accuracy and interpretability, achieving superior results on RAF-DB and SFEW datasets compared to previous YOLO and SSD variants [2].

This direct approach offers the potential to address some limitations of traditional pipelines, such as increased computational cost and error propagation. However, challenges remain in effectively training object detection models for FER, including the need for appropriately labeled datasets and the handling of variations in pose, illumination, and occlusion.

Despite recent advances in Facial Expression Recognition (FER), several gaps persist in the literature. Most studies have focused on older YOLO versions such as YOLOv5 and YOLOv8, with limited investigation into newer, lightweight architectures like YOLOv11 and YOLOv12, which offer promising trade-offs between speed and accuracy. Additionally, FER pipelines commonly follow a multi-stage design that separates face detection from expression classification, resulting in increased complexity and latency. Exploring whether modern YOLO models can unify these tasks in a single-stage framework is crucial for building scalable, real-time systems [2]. Moreover, the adaptability of YOLO for recognizing subtle facial expressions in both constrained and real-world settings remains underutilized. This study aims to address these gaps by adapting and evaluating YOLOv11 and YOLOv12 for FER tasks, using a unified object detection pipeline on FER2013 and KDEF datasets, and assessing their effectiveness across diverse conditions with a focus on resource-efficient deployment.

### III. METHODOLOGY

#### A. Datasets

Two benchmark facial expression recognition (FER) datasets, FER2013 and KDEF, are used in this study. To enable unified training and evaluation across detection-based FER models, both datasets are first converted into the YOLO detection format, where each facial image is annotated with a bounding box and a corresponding emotion label.

**FER2013** contains in-the-wild grayscale images collected from the internet, presenting significant variation in pose, illumination, and occlusion. In contrast, **KDEF** consists of high-resolution, studio-quality color images captured in controlled conditions, offering clean and balanced representations of facial expressions.

Table I summarizes key characteristics of the FER2013 and KDEF datasets, including emotion categories, sample counts, subject demographics, and capture environments. Figure 2 illustrates the distribution of emotion classes in both datasets, highlighting class imbalances. Sample images are shown in Figure 3 to illustrate differences in quality, resolution, and subject presentation.

TABLE I
SUMMARY OF FER2013 AND KDEF DATASETS

| Attribute | FER2013 | KDEF |
|---|---|---|
| Year | 2013 | 1998 |
| Samples | 35,887 | 4,900 |
| Type | Wild | Posed |
| Subjects | Unspecified | 70 (35 Female, 35 Male) |
| Environment | In-the-wild | Studio |
| Expressions | 7 basic | 7 basic |
| Usage | Benchmarking FER in real-world, noisy settings | Benchmarking clean and controlled expressions |
| Limits | Low resolution, class imbalance, no bounding boxes | Limited variation in pose and lighting |
| Link | Download link: [12] | Request link: [9] |

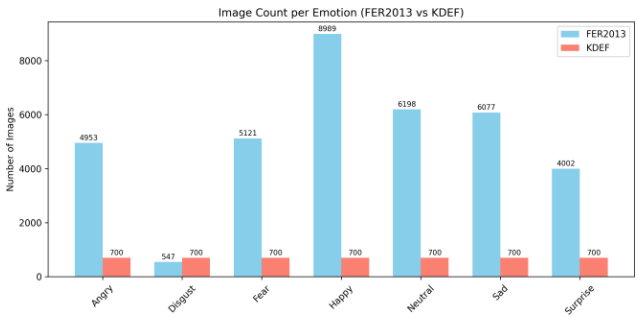

Fig. 2. Distribution of images per emotion in FER2013 and KDEF datasets.

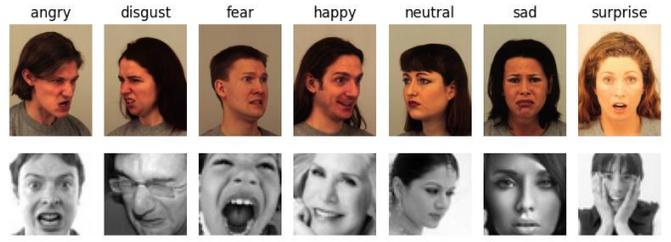

Fig. 3. Example images from the KDEF dataset (top row) and FER2013 dataset (bottom row), illustrating the seven universal facial expressions.

#### B. Dataset Preparation

To enable the use of standard classification datasets in object detection models, the FER2013 and KDEF datasets were converted into the YOLO format. This process involved organizing images into class-specific directories and generating corresponding label files. Both datasets were then split using an 80:20 ratio: FER2013 retained its original split of 28,709 training and 7,178 validation images, while KDEF was manually divided into 3,918 training and 980 validation samples.

Each image was assigned a bounding box covering the entire face, assuming a single, centered subject. For every image, a label file was created containing the class ID and normalized coordinates representing a full-image bounding box (centered at (0.5, 0.5) with width and height of 1.0).

The conversion pipeline was implemented using Python scripts with the `PIL` and `tqdm` libraries. Class names were mapped to integer IDs in the order: *angry*, *disgust*, *fear*, *happy*, *sad*, *surprise*, and *neutral*.

For training and validation, a unified directory structure was adopted with separate folders for images and labels. Additionally, a YOLO-compatible data.yaml file was generated to define the training and validation paths, number of classes, and class names. This format ensures compatibility with the Ultralytics YOLO training pipeline and supports consistent evaluation across datasets.

#### C. Model

This study evaluates two lightweight object detection models: YOLOv11n and YOLOv12n, which are nano versions of the official YOLOv11 and YOLOv12 architectures, respectively. These models belong to the next-generation YOLO series and are optimized for real-time applications with limited computational resources.

YOLOv11 follows a modular design with three main components. Its backbone uses the enhanced C3k2 module for efficient spatial feature extraction. The neck integrates Spatial Pyramid Pooling-Fast (SPPF) for multi-scale feature aggregation, and the head performs detection by predicting bounding boxes and class probabilities in a single-stage pipeline [7].

YOLOv12 builds on this structure by incorporating attention mechanisms in the backbone to better focus on salient facial regions. The neck uses Residual Efficient Layer Aggregation Networks (R-ELAN) for improved feature flow and fusion, while maintaining a streamlined head for accurate, low-latency detection across varied conditions [6].

The nano versions, YOLOv11n and YOLOv12n, are lightweight adaptations of their respective full models, significantly reduce computational overhead while retaining strong detection performance, making them ideal for deployment in low-resource and real-time FER systems.

### D. Model Training

Both YOLOv11n and YOLOv12n models were fine-tuned using transfer learning from their respective COCO-pretrained weights within the Ultralytics YOLO framework. To ensure fair comparison, identical training configurations and hyperparameters were applied across all experiments. The models were trained for 20 epochs with an input image resolution of 320×320 pixels and a batch size of 8. The Adam optimizer was employed with an automatically determined learning rate of 0.000909, momentum of 0.9, and weight decay of 0.0005. Mixed precision training (AMP) was enabled to accelerate computation while maintaining numerical stability.

To enhance model robustness and generalization capability, data augmentation techniques were applied using Albumentations transformations, including horizontal flipping (probability = 0.5), random erasing (probability = 0.4), and Gaussian and median blur (probability = 0.01, kernel size = 3-7). These augmentations help the models adapt to variations in facial orientation, partial occlusions, and image quality differences commonly encountered in real-world facial expression recognition scenarios. The complete set of training hyperparameters is summarized in Table II.

TABLE II
TRAINING HYPERPARAMETERS FOR YOLOv11N AND YOLOv12N

| Hyperparameter | Value |
|---|---|
| Pretrained Weights | COCO |
| Image Size | 320 × 320 |
| Batch Size | 8 |
| Epochs | 20 |
| Optimizer | Adam |
| Learning Rate (auto) | 0.000909 |
| Momentum (auto) | 0.9 |
| Weight Decay | 0.0005 |
| AMP (Mixed Precision) | Enabled |
| Augmentations | FlipLR(0.5), Erasing(0.4), Blur & MedianBlur (p=0.01, limit=3–7) |
| IoU Threshold | 0.7 |
| GFLOPs | 6.3 |
| Workers | 8 |

### E. Evaluation Metrics

Model performance was evaluated using standard object detection metrics: mean Average Precision at IoU threshold 0.5 (mAP@0.5), precision, and recall. In addition, confusion matrices were analyzed to assess per-class prediction accuracy and to identify which facial expressions were most prone to misclassification.

## IV. EXPERIMENTS AND RESULTS

This section presents the experimental setup, evaluation results, and key findings from the performance comparison of YOLOv11n and YOLOv12n for Facial Expression Recognition (FER).

### A. Experimental System Details

All experiments were conducted on Google Colab Pro+ using NVIDIA A100-SXM4-40GB GPU with 40,507 MiB VRAM on a Linux-based backend. The environment was configured with Python 3.11.12, PyTorch 2.6.0+cu124, and the Ultralytics YOLO framework v8.3.146.

### B. Results

Table III summarizes the detection performance of YOLOv11n and YOLOv12n on the FER2013 and KDEF datasets using precision, recall, and mAP@0.5 as evaluation metrics.

TABLE III
DETECTION PERFORMANCE OF YOLOv11N AND YOLOv12N ON FER2013 AND KDEF DATASETS.

| Model | Dataset | Precision% | Recall% | mAP@0.5% |
|---|---|---|---|---|
| YOLOv11n | FER2013 | 65.2 | 60.5 | 60.8 |
| YOLOv11n | KDEF | 87.7 | 91.1 | 94.5 |
| YOLOv12n | FER2013 | 57.3 | 67.1 | 63.8 |
| YOLOv12n | KDEF | 89.6 | 91.2 | 95.6 |

On the KDEF dataset, both models perform strongly due to the clean, studio-quality input. YOLOv12n achieves the highest scores across all metrics, with a mAP@0.5 of 95.6%, recall of 91.2%, and precision of 89.6%, slightly outperforming YOLOv11n. This reflects YOLOv12n's architectural enhancements in feature representation and attention mechanisms, which help in capturing subtle facial cues in clean data.

In contrast, the FER2013 dataset presents greater challenges due to real-world variability in pose, lighting, and occlusion. Here, YOLOv12n demonstrates higher recall (67.1%) and mAP@0.5 (63.8%), suggesting better sensitivity in detecting diverse expressions. However, YOLOv11n outperforms in precision (65.2%), indicating fewer false positives and more reliable predictions in noisy conditions.

The confusion matrices in Figure 4 supports these findings. On the KDEF dataset, both models show strong class-wise performance, with YOLOv12n outperforming YOLOv11n in challenging categories like fear (83% vs. 73%). For expressions such as happy and neutral, both models achieve over 95% accuracy, benefiting from the consistent visual quality of the dataset.

In the FER2013 dataset, the confusion matrices show a notable drop in classification accuracy, particularly for subtle or visually similar expressions like fear, disgust, and sad. YOLOv11n demonstrates better precision on clearly distinguishable expressions such as happy (85%) and surprise (75%), while YOLOv12n captures a higher number of true positives across more difficult classes but suffers from more misclassifications.

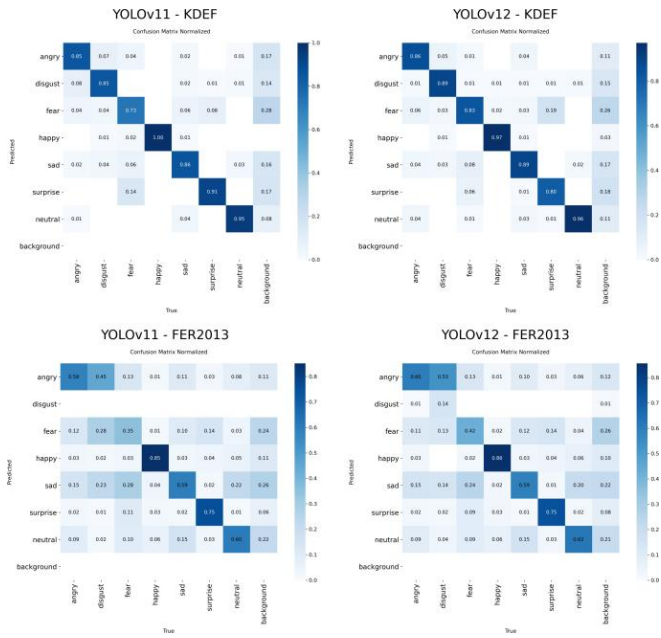

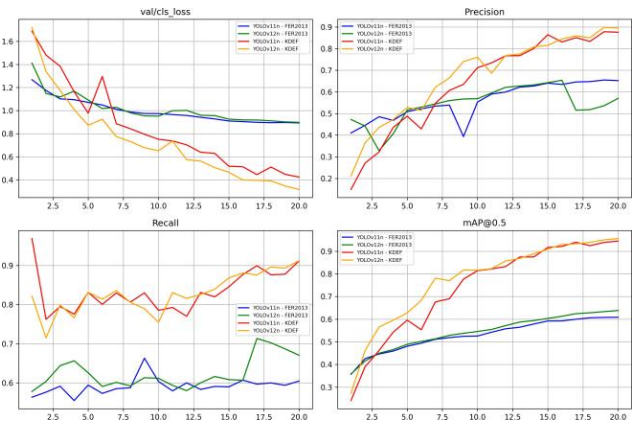

Fig. 5. Training and validation performance comparison across both datasets. Each subplot shows validation loss, precision, recall, and mAP@0.5 over 20 epochs. Curves are color-coded for YOLOv11n and YOLOv12n trained on FER2013 and KDEF.

Fig. 4. Normalized confusion matrices of YOLOv11n and YOLOv12n on KDEF (top) and FER2013 (bottom). Diagonal values represent correct classifications; off-diagonal entries indicate misclassifications, especially among subtle emotions like fear, disgust, and sad.

In YOLO-based detection models, the "Background" class denotes instances where the model fails to assign any facial expression label. In the KDEF dataset, such misclassifications are minimal, remaining below 1% for both YOLOv11n and YOLOv12n due to the dataset's clean, high-resolution, and consistently centered facial images. However, in the FER2013 dataset, background misclassifications are more prevalent, reaching up to 6.3% for YOLOv12n and 4.8% for YOLOv11n. This is mainly due to the presence of low-quality, blurry, or poorly cropped facial images in FER2013, and in some cases, images where the face is hard to see clearly. These missed detections mostly affect subtle emotions like fear, disgust, and sad, which are already difficult to recognize. As a result, the model's ability to correctly detect all expressions (recall) is reduced, showing the difficulty of using FER systems in realworld, unconstrained settings.

Figure 5 presents the training and validation performance of both models. On KDEF, validation loss steadily decreases and all performance metrics converge toward 0.95 for YOLOv12n, indicating excellent detection accuracy. On FER2013, performance trends are noisier. YOLOv11n achieves higher precision (0.65), while YOLOv12n maintains better recall (0.67) and mAP@0.5 (0.64), reflecting a trade-off between detection reliability and expression coverage.

Figure 6 illustrates qualitative prediction examples. Both models perform reliably on clearly visible, frontal facial images from KDEF. Performance degrades on FER2013 images, particularly for expressions with visual similarity (e.g., fear vs. surprise). This figure highlights each model's tendencies and common misclassifications.

### C. Discussion

Overall, YOLOv12n shows superior performance on clean datasets like KDEF, achieving the highest values for precision, recall, and mAP@0.5. On the more challenging FER2013 dataset, YOLOv11n delivers more precise predictions, while YOLOv12n provides broader expression coverage. These results illustrate a trade-off: YOLOv12n is more sensitive but prone to misclassifications, whereas YOLOv11n is more conservative but consistent in its predictions.

Confusion matrices confirm that both models handle distinct expressions (e.g., happy, neutral) with high accuracy, while subtle expressions under noisy conditions remain difficult. YOLOv11n demonstrates better generalization in low-quality settings, while YOLOv12n excels in structured environments, underscoring their complementary strengths.

Despite these promising results, the models have not been tested in more diverse real-world scenarios. Challenges such as partial occlusion, extreme lighting, and varied head poses remain open. Addressing these issues is essential for deploying FER in domains like classroom monitoring, in-vehicle systems, and mental health assessment. Future work may incorporate occlusion-aware training, pose normalization, and domain-specific augmentations to improve robustness. Additionally, evaluating performance on embedded or mobile platforms will help assess real-time usability and efficiency.

## V. CONCLUSION

This study evaluated the performance of lightweight YOLOv11n and YOLOv12n models for facial expression recognition by adapting them to a unified object detection framework. Both models achieved strong results on the clean KDEF dataset, with YOLOv12n delivering the highest accuracy across all metrics. On the more challenging FER2013 dataset, YOLOv11n demonstrated better precision, while YOLOv12n achieved higher recall and mAP, highlighting a trade-off between detection sensitivity and reliability.

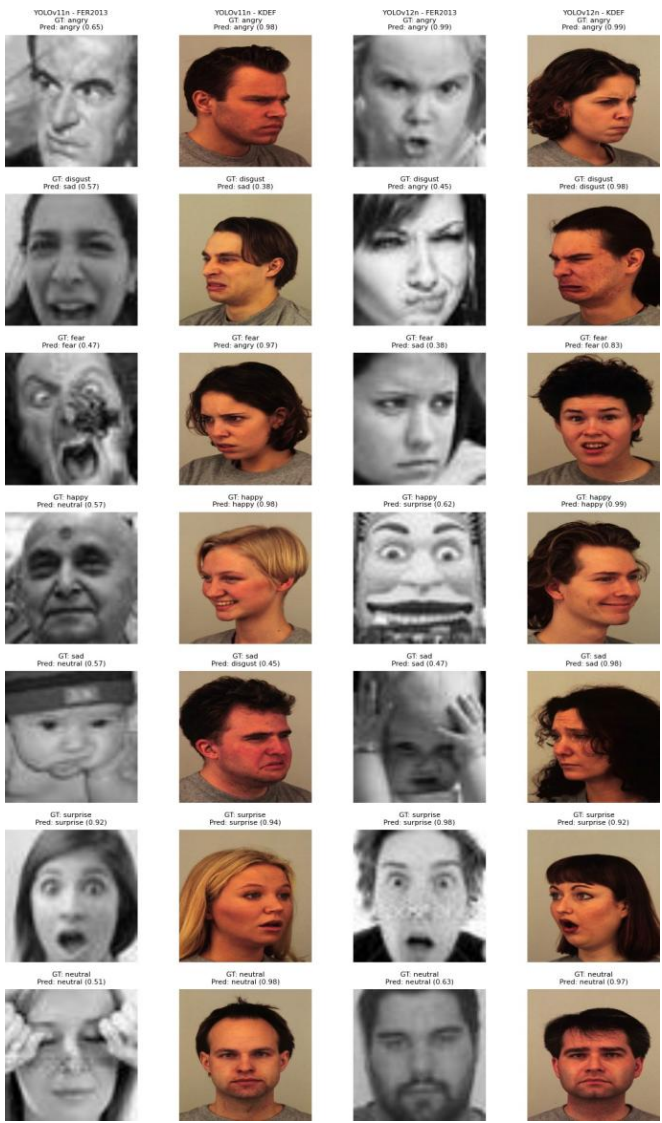

Fig. 6. Visual comparison of expression predictions made by YOLOv11n and YOLOv12n on FER2013 and KDEF datasets. Each row shows one emotion (angry, disgust, fear, happy, sad, surprise, neutral). Columns correspond to model-dataset combinations. Ground truth (GT) and predicted (Pred) labels with confidence scores are shown below each image.

The conversion of classification datasets into YOLO-compatible formats, combined with uniform training procedures, confirms the practicality and effectiveness of using lightweight detection-based models for FER in real-time, low-resource settings. However, further improvements are necessary to enhance model robustness in real-world conditions, particularly when dealing with occlusion, low lighting, and diverse facial poses.

Future work will explore strategies such as occlusion-aware training, pose normalization, and synthetic data augmentation to improve generalization. Additionally, the practical value of these models will be assessed through deployment in real-world applications such as driver monitoring, classroom engagement analysis, or mobile emotion-aware systems.

VI. ACKNOWLEDGMENT

This research has been supported by Malaysia International Scholarship (MIS), Fundamental Research Grant Scheme (FRGS) RDU240125 Tabung Persidangan Dalam Negara and FRGS RDU230117

REFERENCES

[1] M. M. A. Parambil et al., "Navigating the YOLO landscape: A comparative study of object detection models for emotion recognition," IEEE Access, vol. 12, pp. 109427–109442, Jan. 2024, doi: 10.1109/ACCESS.2024.3439346.

[2] H. Ma, S. Lei, T. Celik, and H.-C. Li, "FER-YOLO-Mamba: Facial expression detection and classification based on selective state space," arXiv preprint, arXiv:2405.01828, 2024. [Online]. Available: https://arxiv.org/abs/2405.01828

[3] C. Xu, Y. Du, W. Zheng, T. Li, and Z. Yuan, "Facial expression recognition based on YOLOv8 deep learning in complex scenes," Int. J. Inf. Commun. Technol., vol. 26, no. 1, pp. 89–101, Jan. 2025, doi: 10.1504/IJICT.2025.144013.

[4] M. A. Hasan, "Facial human emotion recognition by using YOLO face detection algorithm," J. Inform., Netw., Comput. Sci. (JOINCS), vol. 6, no. 2, pp. 32–38, 2023.

[5] G. Jocher and J. Qiu, "Ultralytics YOLO11," 2024, GitHub repository. [Online]. Available: https://github.com/ultralytics/ultralytics

[6] Y. Tian, Q. Ye, and D. Doermann, "YOLOv12: Attention-centric real-time object detectors," arXiv preprint, arXiv:2502.12524, 2025. [Online]. Available: https://arxiv.org/abs/2502.12524

[7] M. Chaman, A. E. Maliki, H. El Yanboiy, H. Dahou, H. Laaˆmari, and A. Hadjoudja, "Comparative analysis of deep neural networks YOLOv11 and YOLOv12 for real-time vehicle detection in autonomous vehicles," Int. J. Transp. Dev. Integr., vol. 9, no. 1, pp. 39–48, 2025.

[8] I. J. Goodfellow et al., "Challenges in representation learning: A report on three machine learning contests," in Neural Information Processing: ICONIP 2013, M. Lee, A. Hirose, Z. G. Hou, and R. M. Kil, Eds. Berlin, Germany: Springer, 2013, vol. 8228, Lecture Notes in Computer Science, pp. 117–124. [Online]. Available: https://doi.org/10.1007/978-3-642-42051-1_16

[9] D. Lundqvist, A. Flykt, and A. Öhman, "The Karolinska Directed Emotional Faces (KDEF)," Karolinska Institutet, 1998. [Online]. Available: https://www.kdef.se/. [Accessed: Jun. 27, 2025]

[10] B. Abdellaoui, A. Remaida, Z. Sabri, M. Abdellaoui, A. E. Hafidy, Y. E. B. El Idrissi, and A. Moumen, "Analyzing emotions in online classes: Unveiling insights through topic modeling, statistical analysis, and random walk techniques," Int. J. Cogn. Comput. Eng., vol. 5, pp. 221–236, 2024.

[11] A. Abedi and S. S. Khan, "Engagement measurement based on facial landmarks and spatial-temporal graph convolutional networks," in Proc. Int. Conf. Pattern Recognit., 2024, pp. 321–338.

[12] M. Sambare, "Facial Expression Recognition 2013 (FER2013) Dataset," Kaggle, 2021. [Online]. Available: https://www.kaggle.com/datasets/msambare/fer2013. [Accessed: Jun. 27, 2025]